\documentclass[11pt,letterpaper]{article}
\usepackage{cogsys}
\usepackage[caption=false]{subfig}
\usepackage{url}
\usepackage[T1]{fontenc}
\usepackage{times}
\usepackage[pdftex]{graphicx} 

\usepackage[longnamesfirst]{natbib}
\def\texitem#1{\par\noindent\hangindent 20pt
               \hbox to 20pt {\hss #1 ~}\ignorespaces}
\def\itemitem#1{\par\noindent\hangindent 40pt
                \hbox to 40pt {\hss #1 ~}\ignorespaces}

\cogsysheading{4}{2016}{131-150}{8/2015}{6/2016}

\ShortHeadings{TRESTLE: A Model of Concept Formation in Structured Domains}
              {C.\ J.\ MacLellan, E.\ Harpstead, V.\ Aleven, and K.\ R.\
              Koedinger}

\begin{document} 

\title{TRESTLE: A Model of Concept Formation in Structured Domains}
 
\author{Christopher J. MacLellan}{cmaclell@cs.cmu.edu}
\author{Erik Harpstead}{eharpste@cs.cmu.edu}
\author{Vincent Aleven}{aleven@cs.cmu.edu}
\author{Kenneth R. Koedinger}{koedinger@cmu.edu}
\address{Human-Computer Interaction Institute, Carnegie Mellon University, Pittsburgh, PA 15213 USA}
\vskip 0.2in
 
\begin{abstract}
The literature on concept formation has demonstrated that humans are capable of
learning concepts incrementally, with a variety of attribute types, and in both
supervised and unsupervised settings. Many models of concept formation focus on
a subset of these characteristics, but none account for all of them. In this
paper, we present TRESTLE, an incremental account of probabilistic concept
formation in structured domains that unifies prior concept learning models.
TRESTLE works by creating a hierarchical categorization tree that can be used
to predict missing attribute values and cluster sets of examples into
conceptually meaningful groups. It updates its knowledge by partially matching
novel structures and sorting them into its categorization tree. Finally, the
system supports mixed-data representations, including nominal, numeric,
relational, and component attributes. We evaluate TRESTLE's performance on a
supervised learning task and an unsupervised clustering task. For both tasks,
we compare it to a nonincremental model and to human participants. We find that
this new categorization model is competitive with the nonincremental approach
and more closely approximates human behavior on both tasks. These results serve
as an initial demonstration of TRESTLE's capabilities and show that, by taking
key characteristics of human learning into account, it can better model
behavior than approaches that ignore them.
\end{abstract}

\section{Introduction}

Humans can improve their performance with experience. To better understand
these capabilities, numerous research efforts have constructed computational
models of human learning
\citep{VanLehn:1992ux,Fisher:1990jf,Li:2012tu,Laird:1986ig,Langley:2008ka}.
Early work on human learning embraced categorization as a primary mechanism for
organizing experiences, recalling them in new situations, and using them to
make decisions \citep{Feigenbaum:1984tp,Fisher:1990jf}. In contrast, most recent
work in cognitive architectures emphasizes the generation of solutions to
problems or the execution of actions \citep{Langley:2008ka}. However,
categorization and conceptual understanding remain crucial aspects of
cognition. For example, there is evidence that humans spend more time on
learning to recognize the conditions for an action than on learning the steps
needed to perform it \citep{Zhu:1996ui}. Thus, we argue that more research
should be conducted on human categorization and the role it plays in learning
and problem solving. 

One important aspect of human categorization is that it occurs in an
incremental fashion \citep{GiraudCarrier:2000tq,love2004sustain}. People revise
learned concepts given new information rather than reconsidering all prior
experiences and learning completely new structures. This lets them improve
their performance given more experience and to flexibly adapt their knowledge
and understanding in response to novel situations in an efficient way. This
approach contrasts with the main thrust of research in machine learning, which
emphasizes nonincremental approaches in an effort to aid performance. Thus,
while incremental learners can update their knowledge given new experiences,
their performance is affected by the order in which these experiences occur.
For example, when humans are learning to solve fractions problems, the order in
which they receive problems affects their learning \citep{Rau:2013gc}. More
generally, how best to order practice problems for humans in order to
promote robust learning remains an open research question
\citep{Li:2012ex,Carvalho:2013kt} that would stand to benefit from computational
models of incremental acquisition.

A second characteristic of human categorization is that it occurs in a wide
range of environments that can involve structural, relational, nominal, and
numeric information. For example, the work of \citet{Carvalho:2013kt} has
shown that the structural similarity of objects present in examples affects
human concept formation. Other work has identified structure mapping as a
crucial ability that allows humans to map their prior concepts to new
situations \citep{Forbus:1995tr,Holyoak:2005wi}. While prior models of concept
learning have explored how to handle relational \citep{Quinlan:1995uc}, nominal
\citep{Fisher:1987tb}, and numeric
\citep{Gennari:2002tl,Li:2002ut,Quinlan:1986cv} information, less emphasis has
been placed on learning with structural content
\citep{Thompson:2010wv,Forbus:1995tr}. Further, few studies exist that explore
how best to model the integration of multiple data types. Of the existing
methods, only a subset integrate nominal and numeric data
\citep{Li:2002ut,Quinlan:1986cv,Gennari:2002tl}. Finally, even fewer studies
combine relational \citep{Quinlan:1995uc} or structural
\citep{Thompson:2010wv,Forbus:1995tr} information. We claim that any system
attempting to model human categorization should integrate all of these data
types into a single account. 

A third characteristic of human concept learning is that it occurs in both
supervised and unsupervised settings \citep{love2004sustain}. In many cases,
these two modes of learning are treated as distinct, but there is evidence that
humans can improve their performance on unsupervised tasks given supervision on
a different task \citep{zhu2007humans}. Others have found the reverse, that
humans can improve their performance on supervised tasks given experience on
other unsupervised tasks \citep{Kellman:2009ji}. Taken together, these findings
suggest that humans share knowledge across these settings, and that models of
human categorization should be capable of operating with and without
supervision, with knowledge used for one influencing the other. 
 
To explore these aspects of human categorization, we developed TRESTLE, an
incremental model of probabilistic concept formation in structured domains that
builds on prior research in categorization
\citep{Fisher:1987tb,McKusick:2013vm,Thompson:2010wv}. This approach maps novel
structures to its existing knowledge in an online fashion and updates its
categorization knowledge based on these new structures. TRESTLE also handles
mixed representations, letting it function with nominal, numeric, relational,
and component attributes. Finally, the approach supports partial matching, so
it can process incomplete or partially specified input. These features let
TRESTLE use its categorization knowledge for predicting missing attributes in
supervised settings and for clustering examples in purely unsupervised
situations. In this paper, we present an example domain based on the
educational game {\it RumbleBlocks} \citep{Christel:2012dr} that contains
nominal, numeric, relational, and structured information, describe TRESTLE's
approach to learning concepts in this domain, and present a preliminary
evaluation of the algorithm that compares it with more specialized approaches
and with humans. We show that the model can learn probabilistic concepts given
supervision and make predictions based on these concepts at human levels of
performance. Additionally, for unsupervised clustering, it produces clusters
that have reasonably high agreement with human clusterings. These results
provide an initial demonstration of how TRESTLE models human concept formation
in structured domains. 

\section{The {\it RumbleBlocks} Domain}

\begin{figure}[t!]
  \centering
    \includegraphics[width=0.5\textwidth]{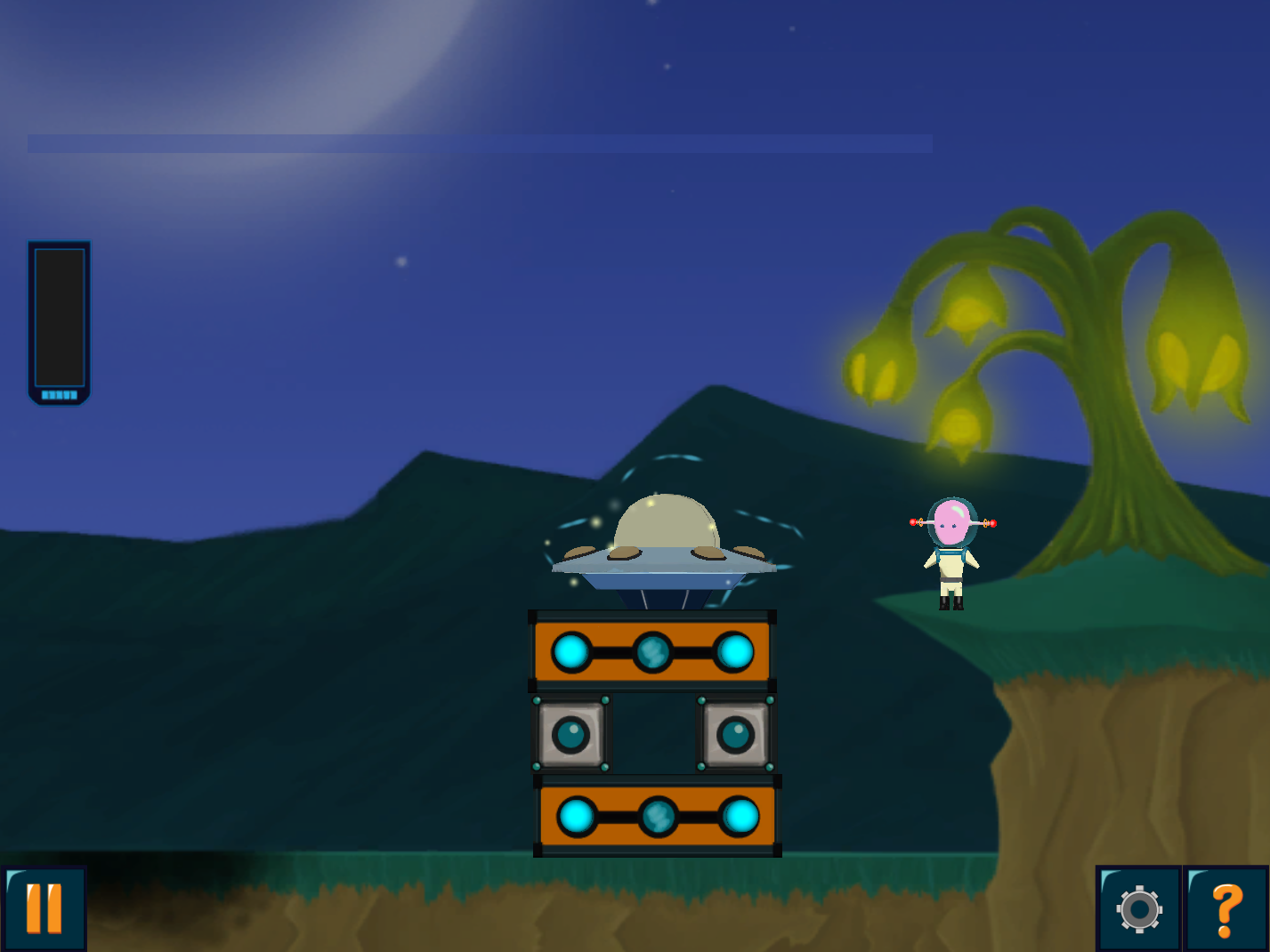}
    \caption{A screenshot of a player building a tower in {\it RumbleBlocks}.
    The final tower must cover the light blue energy balls and survive a
    simulated earthquake to be successful.}
  \label{fig:rbscreen}
\end{figure}

To investigate human learning in a rich domain, we have explored data from {\it
RumbleBlocks} \citep{Christel:2012dr}, an educational game designed to teach
concepts of structural stability and balance to young children. What makes this
domain interesting is the detailed structural quality of game tasks.  Players
build a tower out of blocks in a two-dimensional, continuous world with a
realistic rigid body physics simulation, designing their towers to cover a
series of energy balls in an attempt to power an alien's spaceship (see Figure
\ref{fig:rbscreen}). These energy balls function as both scaffolding and
constraints on players' designs. After designing their tower, players place the
spaceship on the tower, which charges the ship and causes an earthquake. If the
tower survives the earthquake with the spaceship intact, then the player
succeeds and goes on to the next level; otherwise he fails and must
try the level again.

Each level of the game is designed to emphasize one of three primary concepts
of structural stability and balance: objects that are symmetrical, that have 
wide bases, and that have lower centers of mass are more stable. While the
drag-and-drop actions of tower construction are necessary to succeed in the
game, they are not relevant its educational goals. The central learning
challenge for players of {\it RumbleBlocks} is to understand how to categorize
the states of the game world as good or bad examples of the game's target
concepts.

{\it RumbleBlocks} provides several interesting hurdles when attempting to
model players' conceptual learning. The biggest challenge is in representing
the richly structured space of the game. States exist in a continuous space but
also possess nominal (e.g., block types) and structural information (e.g., high
level patterns like arches). Additionally, the physics simulation that models
the earthquake is nondeterministic and susceptible to minor perturbations in
rigid body physics. Therefore, two towers that appear similar have a small
chance of being treated differently in terms of the success criteria, resulting
in noisy evaluation feedback. This noisy feedback makes it more challenging to
learn the correct concepts and makes ordering effects more pronounced; if a
borderline tower fails early in a student's problem sequence, then he will be
more likely to avoid similar structures throughout his play than if the tower
had succeeded. Finally, {\it RumbleBlocks} lets students partake in both
supervised and unsupervised learning. They receive supervised feedback
regarding the success of their towers, but are left to learn the game's target
concepts (i.e., wide base, low center of mass, and symmetry) in an unsupervised
way.

An early attempt exploring how players learn in {\it RumbleBlocks} sought to
model the acquisition of structural patterns as a grammar induction task
through a process called CFE or Conceptual Feature Extraction
\citep{Harpstead:2013vja}.  CFE, which is similar to the grammar learning
approach used by SimStudent \citep{Li:2012vz},  takes a series of examples and
discretizes them to a grid before exhaustively generating a two-dimensional
context-free grammar that is capable of parsing each example in every possible
way. The parse trees for each example are then converted into feature vectors
using a bag of words approach, where each symbol in the grammar has its own
feature. The generated feature vectors are then used by traditional learning
methods for prediction or clustering. The goal was to construct features that
capture the maximal amount of structural information in the examples so that
this information could be used in learning. 

The CFE approach was successful for clustering student solutions in order to
understand how players' design patterns differed from those expected by the
game's designers \citep{Harpstead:2013vja}, but it has not been evaluated as
part of a model of human categorization in {\it RumbleBlocks}. There are
several obstacles to using CFE to model human categorization. First, the
algorithm runs in a batch fashion instead of incrementally and requires all the
examples for each task to generate its grammar. Thus, any examples with
previously unseen features would require the learning of an entirely new
grammar. Second, the grammar rule formalism used by CFE breaks down in
continuous environments and when two-dimensional objects cannot be cleanly
decomposed across axes. For example, when given a set of blocks in a spiral
pattern, it cannot decompose the structure along the horizontal or vertical
axes; subsequently, it cannot reduce the structure to the primitive grammar
elements and fails at parsing the block structure. Finally, CFE and other
similar grammar approaches \citep{Talton:2012wz} cannot handle missing data in
that they cannot partially match during parsing. Thus, if an attribute is
missing from the training set, then examples containing that attribute
cannot be parsed during run time.  Similarly, if an attribute is always present
in training but not at run time, then some examples would not be parsable. 

\section{The TRESTLE Model}

In order to better model human concept formation in the {\it RumbleBlocks}
domain we developed TRESTLE,\footnote{TRESTLE source code is available at
\url{http://github.com/cmaclell/concept_formation}} a system that incrementally
constructs a categorization tree given the sequential presentation of
instances. This categorization tree can then be used to make predictions about
the given instances or to generate cluster labelings. At a high level, TRESTLE
proceeds through three major steps when given an instance: 

\begin{enumerate}
\setlength{\itemsep}{1pt}
\setlength{\parskip}{0pt}
\setlength{\parsep}{0pt}
\item {\bf Partial matching}, which structurally renames the instance to align it with existing concepts;
\item {\bf Flattening}, which converts a structured instance into an unstructured one while preserving its structural information with a specific naming scheme; and
\item {\bf Categorization}, which incorporates the example into existing knowledge and, if necessary, makes predictions. 
\end{enumerate}

\noindent In this section we describe how TRESTLE represents both
instances and concepts. We also clarify the processing steps that it carries out to learn from each instance.

\subsection{Instance and Concept Representations}

\begin{figure}[t!]
  \centering
    \includegraphics[width=\textwidth]{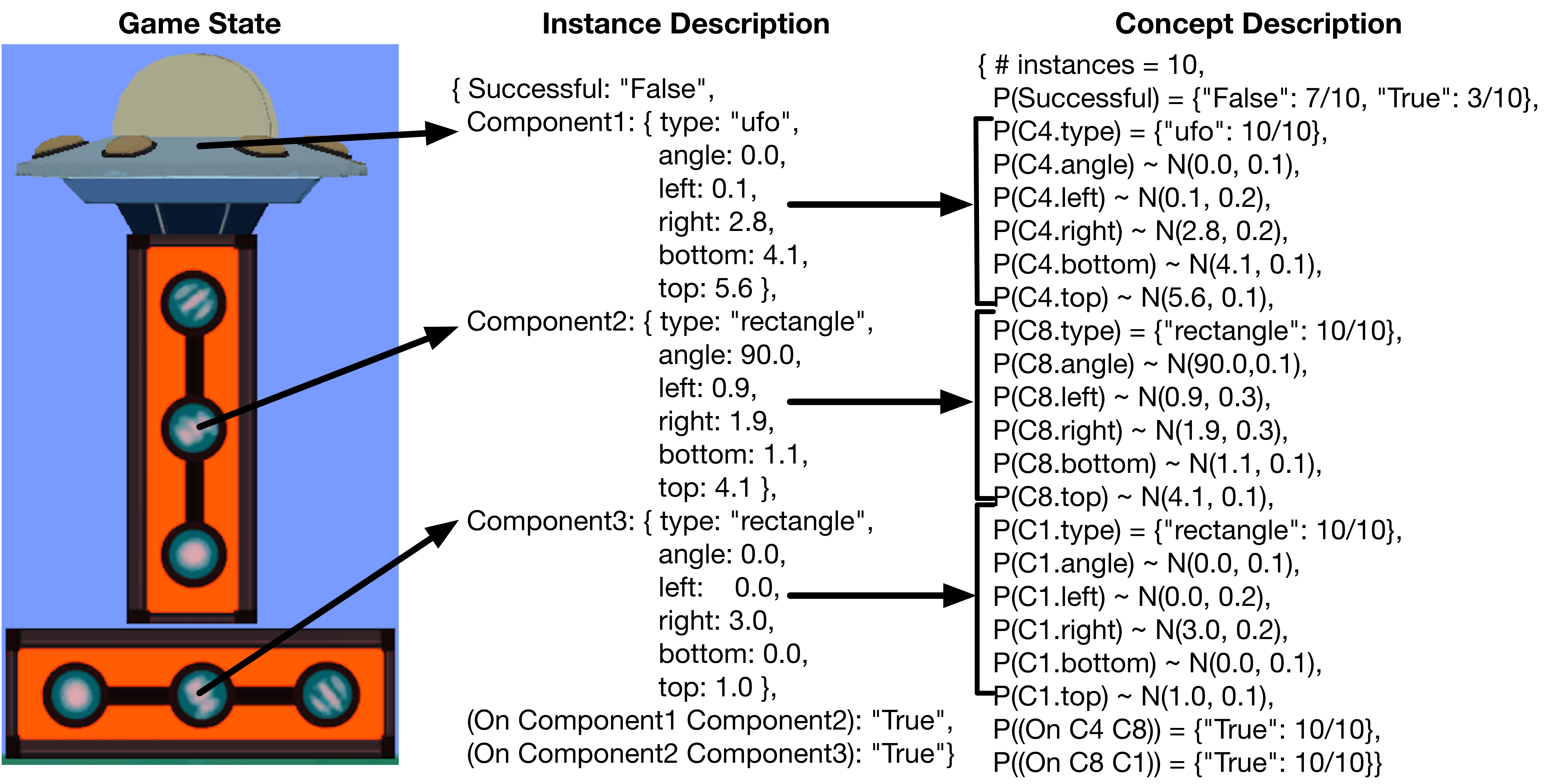}
  \caption{A tower in {\it RumbleBlocks}, its representation as an instance in
  TRESTLE using the four attribute-value types (nominal, numeric, component,
  and relational), and the representation of a TRESTLE concept that might
  describe the instance. The concept stores the number of instances categorized
  as this concept, the probability of each nominal attribute value given their
  occurrence counts, and the normal density function for each numeric attribute
  given the mean and standard deviation of their values. The arrows denote the mapping between blocks, instance components, and components in the concept.}
  \label{fig:av-example}
\end{figure}

TRESTLE uses two representations to support learning: one for instances (i.e.,
specific examples it encounters) and one for concepts (i.e., generalizations of
examples in memory). Instances are encoded in TRESTLE as sets of attribute
values, where each attribute can have one of four types: {\bf nominal} (e.g.,
the type or color of a block); {\bf numeric} (e.g., the position of a block in
continuous space); {\bf component} (e.g., a block with type and dimension
attributes is a component of a greater tower); and {\bf relational} (e.g., that
two blocks are vertically adjacent). Figure \ref{fig:av-example} shows an
example of a {\it RumbleBlocks} tower and how it is represented using these
four attribute-value types. In this example, we include the ``On'' relation to
demonstrate how relations are encoded, but in practice the component
information is often sufficient for learning meaningful concepts. Our
subsequent evaluation does not use these hand-generated relations; we only used
the component and success information that was automatically provided by the
game engine.

In response to being presented with instances, TRESTLE forms a hierarchy of
concepts, where each concept is a probabilistic description of the collection
of instances stored under it. This probabilistic description is stored in the
form of a probability table that tracks how often
each attribute value occurs in the underlying instances (e.g., see Figure
\ref{fig:av-example}). These probability tables can be used to lookup the
probability of different attribute values occurring in an instance given its
concept label. Additionally, the concept maintains a count of the number of
instances it contains, so that the probability of the concept given its parent
concept can be computed. 

\subsection{Partial Matching}

When presented with an instance, TRESTLE searches for the best partial match
between the instance and the root concept, which contains the attribute-value
counts of all previous instances. Each match consists of a partial mapping
between the component attributes in the instance to component attributes in the
root concept. For example, Component1 in the instance shown in Figure
\ref{fig:av-example} might be renamed to C4. When component attributes are
renamed, any relation attributes that reference them are also updated; for
instance (On Component1 Component2) might become (On C4 C8). TRESTLE currently
only performs matching at the root for efficiency reasons, although future work
might explore effects of matching on intermediate concepts.

The quality of a match is determined by how similar the resulting instance is
to the root concept. Similarity is computed as the expected number of attribute
values that can be correctly guessed after incorporating the matched instance
into the root. Under the assumption that an attribute $A_i$ with value
$V_{ij}$ can be guessed with the probability $P(A_i = V_{ij})$ and is correct
with the same probability, maximizing the similarity is equivalent to
maximizing $\sum_i \sum_j P(A_i = V_{ij})^2$. This optimization function can
be efficiently computed using only the root concept's probability table.

To select the best match, TRESTLE searches the space of all matches using
best-first search. To heuristically evaluate the quality of each state in this
search TRESTLE computes the change in the number of expected correct guesses
for the component attributes it has already matched as well as the best
possible improvement in expected correct guesses for unmatched component
attributes. By default, TRESTLE uses this heuristic to guide a beam search
\citep{Wilt:2010wk} under the assumption that greedy approaches are more
psychologically plausible than optimal ones. However, this heuristic is
admissible and ensures that the best possible match is found when used with the
A* algorithm, which TRESTLE optionally supports. The mapping between the
instance and concept in Figure \ref{fig:av-example} is an example of a best
match using beam search. 

\subsection{Flattening}

After matching the instance to the root concept, TRESTLE then flattens the
instance using two procedures. First, it converts all relational attributes
directly into nominal attributes. Second, component attributes are eliminated
by concatenating component and attribute names into a single attribute name.
Throughout the paper we denote this conventionally using dot notation. For
examples, the instance in Figure \ref{fig:av-example} would be flattened as
\{Successful: ``False'', Component1.type: ``UFO'',  Component1.angle: 0.0,
Component1.left: 0.1, Component1.right: 2.8, $\cdots$, ``(On Component1
Component2)'': ``True'', ``(On Component2 Component3)'': ``True'' \}. The
flattening process effectively eliminates the component and relational
attributes while preserving their information in a form that can be used during
partial matching to rename later instances. Flat representations can be
converted back into a form that contains component and relational attributes.
Once converted into a flat representation the instances only contain nominal
and numeric attributes that can be handled by existing approaches to
incremental categorization.

\subsection{Categorization}

\begin{figure}[t!]
  \centering
    \includegraphics{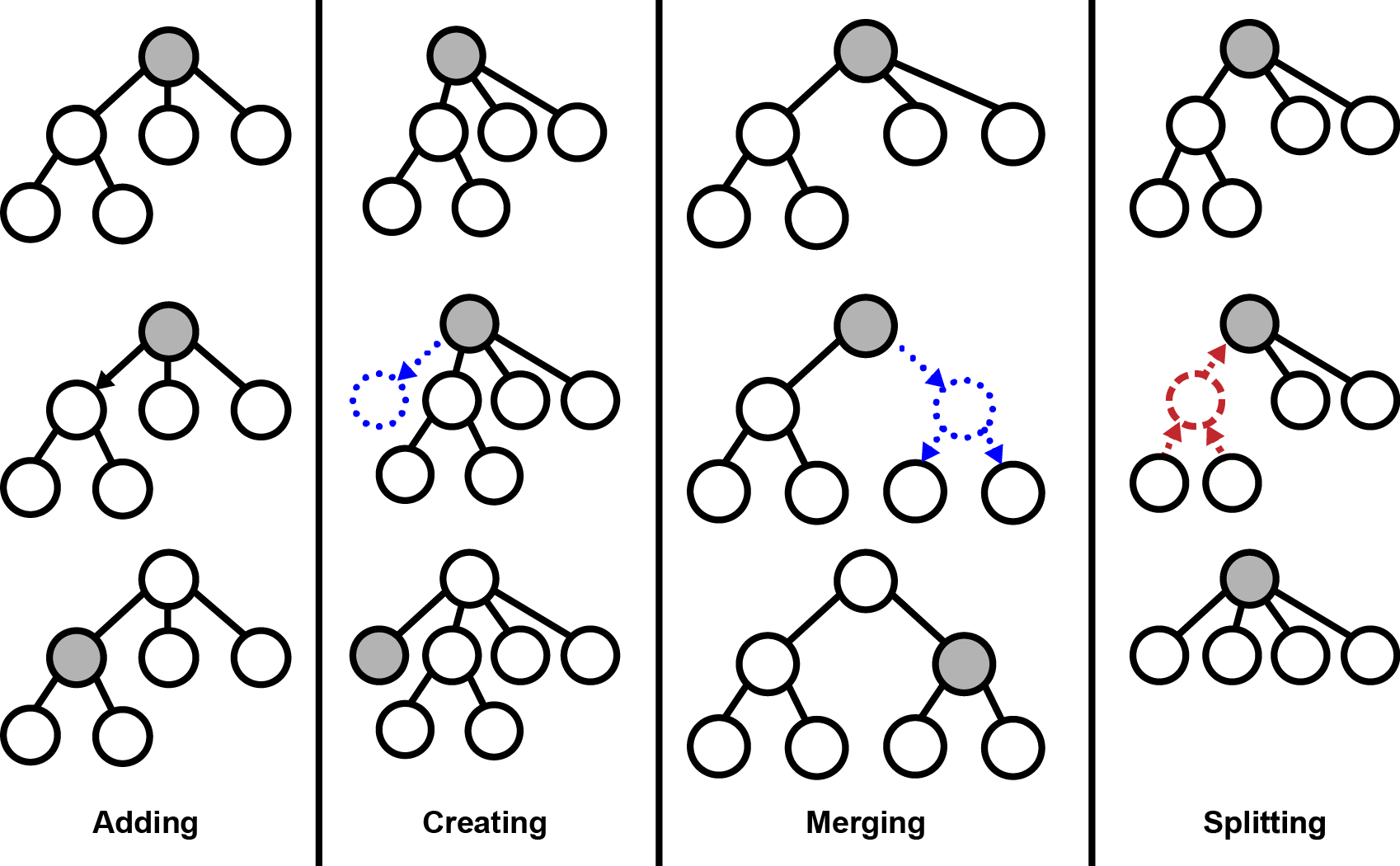}
  \caption{The four operations used by the COBWEB \protect\citep{Fisher:1987tb}
  to incorporate matched instances into its categorization tree. Each shaded
  node depicts the location of the instance being sorted into the tree before
  and after an operation. The blue dotted lines represent nodes and links
  that are being added to the tree and red dashed lines represent nodes and
  links that are being removed from the tree.}
  \label{fig:operations}
\end{figure}

To categorize flattened instances, TRESTLE employs the COBWEB algorithm
\citep{Fisher:1987tb,McKusick:2013vm}, which recursively sorts instances into a
hierarchical categorization tree. At each concept encountered during sorting,
it considers four possible operations (shown in Figure \ref{fig:operations}) to
incorporate the instance into its tree: {\bf adding} the instance to the most
similar child concept; {\bf creating} a new child concept to store the
instance; {\bf merging} the two most similar child concepts and then adding the
instance to the resulting concept; and {\bf splitting} the most similar child
concept, promoting its children to be children of the current concept, and
recursing. COBWEB determines which operation to perform by simulating each
action and computing the category utility \citep{Fisher:1987tb} of the resulting
child concepts. Like the similarity value being maximized during partial
matching, this represents the increase in the average number of expected
correct guesses achieved in the children compared to their parent concept; thus
it is similar to the information gain heuristic used in decision-tree induction
\citep{Quinlan:1986cv}.  Mathematically, the category utility of a set of
children $\{C_1, C_2, \cdots, C_n\}$ is

\begin{eqnarray*}
CU(\{C_1, C_2, \cdots, C_n\}) &=& \frac{\sum_{k=1}^{n} P(C_k)[\sum_i
\sum_j P(A_i = V_{ij} | C_k)^2 - \sum_i \sum_j P(A_i =
V_{ij})^2]}{n}\hspace{4pt},
\end{eqnarray*}

\noindent where $P(C_k)$ is the probability of a particular concept given
its parent, $P(A_i = V_{ij}| C_k)$ is the probability of attribute $A_i$ having
value $V_{ij}$ in the child concept $C_k$, $P(A_i = V_{ij})$ is the probability
of attribute $A_i$ having value $V_{ij}$ in the parent concept, and $n$ is the
number of child concepts. Each of these terms can be efficiently computed
via a lookup of the probability tables stored in the parent and child concepts. 

For numeric attributes, COBWEB uses a normal probability
density function to encode the probability of different values of a numeric
attribute. Given this assumption, the sum of squared attribute-value
probabilities is replaced with an integral of the squared probability density
function, which in the case of a normal distribution is simply the square of
the distribution's normalizing constant. Thus, $\sum_i \sum_j P(A_i = V_{ij} |
C_k)^2 - \sum_i \sum_j P(A_i = V_{ij})^2$ is replaced with $\sum_i
\frac{1}{\sigma_{ik}} - \frac{1}{\sigma_i}$ in the case of numeric attributes,
where $\sigma_{ik}$ is the standard deviation of values for the attribute $A_i$
in child concept $C_k$ and $\sigma_{i}$ is the standard deviation of values for
the attribute $A_i$ in the parent concept. For more detailed justification of
this modification to category utility, see \citet{Gennari:2002tl}. 

\begin{figure*}[t!]
    \centering
    \subfloat[Original tree\label{fig:simple-a}]{
        \centering
        \includegraphics[height=2.2in]{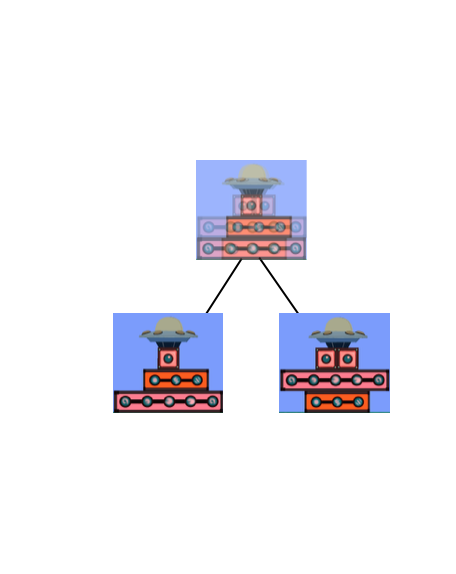}
    }
    ~ 
    \subfloat[Creating a new concept\label{fig:simple-b}]{
        \centering
        \includegraphics[height=2.2in]{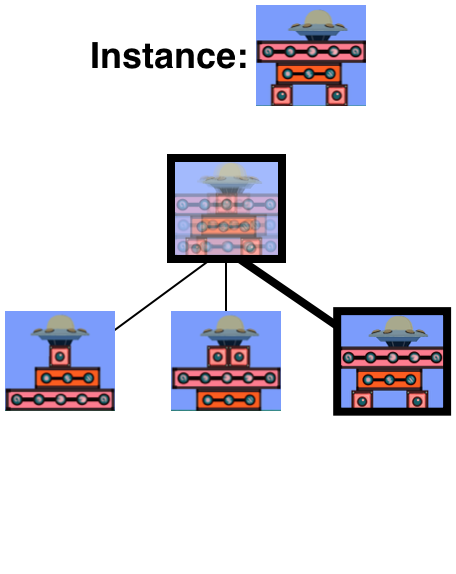}
    }
    ~ 
    \subfloat[Merging two existing concepts and creating a new one\label{fig:simple-c}]{
        \centering
        \includegraphics[height=2.2in]{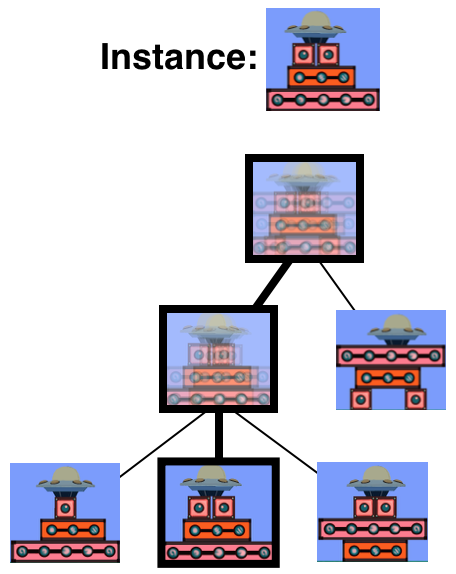}
    }
    \caption{A simple example of how TRESTLE's categorization tree is updated
    in response to two new instances. The original tree (a) is modified in (b)
    and (c) to incorporate the instances shown at the top. In each case, the
    path of the instance through the categorization tree is shown in bold. The
    concepts are depicted as overlapping images of the instances that they
    contain to represent their probabilistic nature.}
\label{fig:simple-example}
\end{figure*}

Figure \ref{fig:simple-example} shows a simple example of a TRESTLE
categorization tree from the {\it RumbleBlocks} domain and how it is updated in
response to new instances. For the purposes of the example, we represent
concepts as the images of the instances they contain rather than probability
tables. Thus, concepts with a single instance are a single image while concepts
with many instances are represented as multiple overlapping images to reflect
their probabilistic nature. Figure
\ref{fig:simple-example}\subref{fig:simple-a} shows an existing categorization
tree containing two previously incorporated instances. In Figure
\ref{fig:simple-example}\subref{fig:simple-b}, TRESTLE is presented with a new
instance, evaluates the category utility of its four operations, and decides
that {\bf creating} a concept to represent the new instance has the highest
utility. The new concept is shown in the right leaf of Figure
\ref{fig:simple-example}\subref{fig:simple-b}. In Figure
\ref{fig:simple-example}\subref{fig:simple-c}, TRESTLE is presented with
another example. In this case, the new instance is similar to two existing
concepts, so TRESTLE decides that merging the two concepts and adding the new
case to the merged concept has the highest category utility. After performing
the {\bf merge} action, the system continues categorization by considering the
four operations at the merged node and decides that {\bf creating} a new
concept to represent the instance has the highest utility. The newly created
concept is depicted by the middle node at the bottom of the tree in Figure
\ref{fig:simple-example}\subref{fig:simple-c}. When TRESTLE encounters future
instance, it updates the tree in a similar fashion using the four operations
shown in Figure \ref{fig:operations}. 

\subsection{Prediction and Clustering}

During learning, COBWEB uses this categorization technique to incorporate new
instances into its conceptual hierarchy. The resulting tree can then be used to
make predictions about novel instances or provide clusterings of the examples
to varying depths of specificity. 

When TRESTLE makes predictions, it follows its normal partial matching and
flattening procedures but employs COBWEB's non-modifying categorization
process. In this version of categorization, an instance is sorted down the
tree, but concepts do not update their probability tables to reflect the
example.  Additionally, the system only considers the {\bf creating} and {\bf
adding} operations at each concept.  When categorization encounters a leaf or a
situation where the {\bf creating} operation has the highest category utility,
it returns the current concept. The resulting concept's probability table is
then used to make predictions about attribute values of the instance, with each
missing attribute predicted to have its most likely value according to the
table.

In addition to prediction, TRESTLE can cluster data by using COBWEB to
categorize instances into its concept tree and assign labels based on the
selected concepts. Using this process, TRESTLE produces a hierarchical
clustering of the data. Alternatively, TRESTLE can convert this hierarchical
clustering into a flat clustering by labeling all instance by the most general
unsplit label (initially the root concept for all instances). For more
specific cluster labels, TRESTLE progressively applies COBWEB's {\bf splitting}
procedure to the most general unsplit concept labels until a desired level of
specificity is achieved.

\section{Experimental Evaluation}

We designed TRESTLE to model three key aspects of human concept learning:
incremental processing, use of multiple information types (nominal, numeric,
component, and relational), and operation in both supervised and unsupervised
settings. Further, implicit in this design is the claim that, by taking these
three aspects of human categorization into account, we can better model human
categorization. Thus, we had two goals for our experimental evaluation: to
demonstrate TRESTLE's abilities to perform incremental learning with multiple
types of information in both supervised and unsupervised capacities, and to
show that, by taking these into account, it models human performance better
than similar systems that do not. 

To demonstrate TRESTLE's functionality, we assessed its behavior on two tasks:
a supervised learning task and an unsupervised clustering task. For each task
we used player log data from the {\it RumbleBlocks} domain, which contains
nominal, numeric, and component information. Additionally, we wanted to test
our claim that the system models human categorization better than alternative
ones. To achieve this goal, we compared TRESTLE to CFE, a nonincremental
approach to learning in structured domains. To establish a baseline for the
comparison, we had humans perform both tasks. We then compared the behavior of
both systems to the human behavior to assess which approach better models
the latter's learning.

\subsection{Task 1: Supervised Learning}

For the supervised learning task, each learner (TRESTLE, CFE, and human) was 
sequentially presented with {\it RumbleBlocks} towers and asked to predict if
the tower successfully withstood the earthquake. They were then given
correctness feedback about their prediction (i.e., provided the success label).
Instances for this task were taken randomly from all player solutions to the
same level of {\it RumbleBlocks}. To avoid a naive strategy of base rate
prediction, we chose a level whose overall success ratio was roughly even (a
symmetry level in which 48.9\% of player towers stood). We presented each
learner with 30 randomly selected and randomly ordered examples and averaged
the accuracy for each approach across opportunities.

\begin{figure}[t!]
  \centering
    \includegraphics[width=0.4\textwidth]{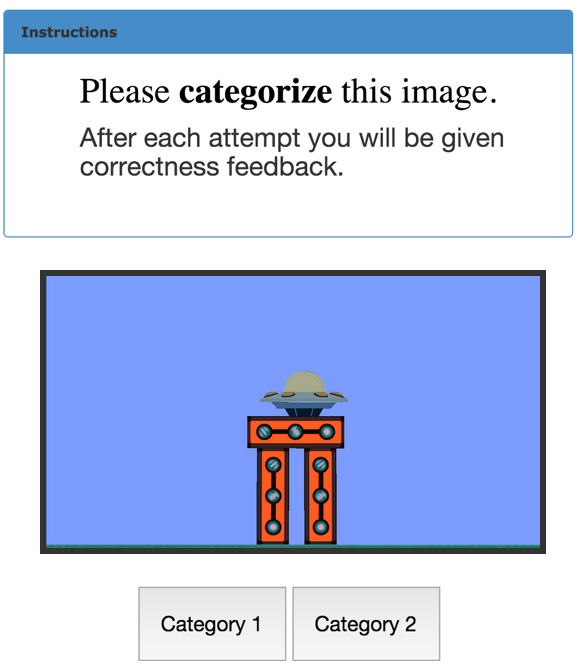}
  \caption{The sequential prediction task as presented to Amazon Mechanical
  Turk workers, who were asked to categorize 30 {\it RumbleBlocks}
  solutions into one of two categories: Category 1 (successful) or Category 2
  (unsuccessful). They were not given any information about the meaning of the
  category labels, but they were given correctness feedback after each
  attempt.}
  \label{fig:turk}
\end{figure}

To determine human behavior for this task, we used Amazon
Mechanical Turk to have 20 humans perform sequential prediction using the
interface shown in Figure \ref{fig:turk}. We asked participants to decide if a
screenshot of an example tower fit into ``Category 1'' (successful) or
``Category 2'' (unsuccessful) before telling them the correct category. We
presented labels abstractly to avoid human raters using their intuitive physics
knowledge, which would not be accessible to either of the computational models.

CFE is a nonincremental approach for handling structural information that must
be paired with another learning algorithm in order to do prediction or
clustering. To sequentially predict the success labeling of solutions, we
paired CFE with CART, a decision-tree learner implemented in the Scikit-learn
library \citep{scikit-learn}. Because CART is nonincremental, we rebuilt its
decision tree after viewing each new training instance. We applied this
approach to 1,000 random training sequences and averaged the correctness of the
predictions across opportunity counts. We used CART for prediction because its
internal structure is similar to TRESTLE's (i.e., they both construct tree
structures), but learning in CART is nonincremental and optimizes for
prediction of a single attribute. This similarity on all but these two
dimensions makes CART a reasonable candidate for testing the importance of
incrementality and multi-attribute prediction when modeling human
categorization. 

We applied TRESTLE to this task by having it predict each instance's success
label. After each prediction, we trained it with the correct instance label in
order to update its knowledge base. Similar to our approach with CFE, we
performed this process with 1,000 random training sequences and averaged the
correctness of the predictions across opportunity counts.

\begin{figure}[t!]
  \centering
    \includegraphics[width=0.9\textwidth]{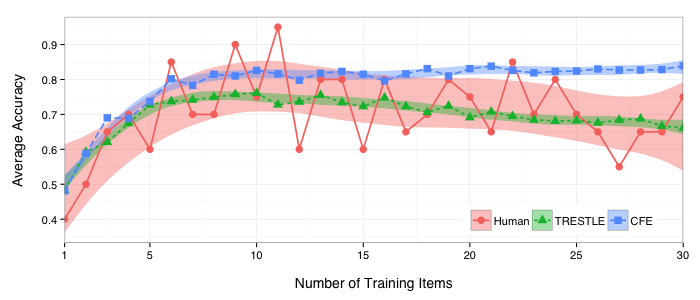}
  \caption{The average performance of 20 participants, 1,000 runs of TRESTLE,
      and 1,000 runs of CFE for predicting whether solutions to a {\it
      RumbleBlocks} level survived an earthquake (shaded regions denote the
      95\% Lowess confidence intervals). We see that the incremental TRESTLE
  algorithm performs less well than the nonincremental CFE algorithm, but that
  TRESTLE more closely approximates human performance.}
  \label{fig:prediction}
\end{figure}

The results of average prediction accuracy for 30 sequentially presented
instances can be seen in Figure \ref{fig:prediction}. Human labelers appear to
converge around 70\% accuracy, suggesting that learning the success concept is
somewhat difficult for them, likely due to both the abstraction
presented in the interface and the noise in the outcome variable. In contrast,
the CFE approach performs better than humans, converging to roughly 83\%
accuracy. TRESTLE performs roughly equal to the humans, with a difference in
confidence intervals that is most likely due to the differences in sample size
(20 humans vs. 1,000 agents).

CFE learns more rapidly, with a steeper learning curve, probably because it is
nonincremental, using a previously generated grammar, and processing all
training cases at each step in batch fashion. Its higher asymptotic accuracy
most likely results from optimizing for prediction of a single attribute. In
contrast, TRESTLE learns incrementally from the examples producing a shallower
learning curve and optimizes for its ability to predict all attributes, so its
overall accuracy predicting the single attribute of success is lower. Our
results suggest that, although these differences decrease predictive accuracy,
they result in a model that more closely aligns with human behavior. This
agrees with previous studies on human category learning, which suggest that
humans often optimize for more general and flexible learning goals, and that
this optimization occurs incrementally \citep{love2004sustain}. Overall, these
findings support our claim that TRESTLE is a better model of human learning
than CFE because it is incremental and it commits to being able to predict any
missing attribute value. CFE may function better as a pure machine learning
system and achieve higher accuracy, but it is a poorer model of human learners
on the RumbleBlocks task than TRESTLE. 

\subsection{Task 2: Unsupervised Clustering}

The results of the supervised learning task show that TRESTLE performs
similarly to humans, but we are also interested in demonstrating its
unsupervised learning capabilities and investigating whether its underlying
representation of knowledge is qualitatively similar to that in humans. A
strong similarity between their organizations of knowledge would strengthen our
claim that it offers a better model of human learning than CFE. In particular,
if its organization of concepts aligns with that of humans, it would suggest
that the agreement with humans on the supervised learning task is the result of
a similar organization of knowledge rather than both learners performing poorly
by chance. To examine this aspect of TRESTLE, we used an unsupervised learning
task in which towers were clustered without any information about their success
or other category labels. For this evaluation, we chose records from three
levels of {\it RumbleBlocks} that were part of an in-game test. These levels
were attractive because they have many player solutions and because the energy
ball mechanic was removed for the test setting, allowing for a wider variance
of player solutions than other levels.

To establish a human baseline, we had two researchers from our laboratory
independently hand cluster screenshots of all the player solutions to each
level. These raters were told to group solutions that looked similar and to
ignore any blocks that did not appear to be part of the solution tower (e.g.,
blocks off to the side of the frame), but were not given any other guidance. To
measure agreement between raters, we calculated the adjusted Rand index between
their clusterings \citep{rand1971}. This measure is a generalization of Cohen's
Kappa that allows for raters to use different numbers of categories.  Values
range from -1 to 1, with random clusterings producing an average of zero. The
average obtained across all three levels was 0.88, which is high. Given the
reasonably high agreement between raters, we chose one of the human clusterings
to be used in further comparisons with the machine approaches. The labeling
from this rater had an average of 33 clusters across the three levels.

As a comparison, we applied the CFE process to generate feature vectors
for the {\it RumbleBlocks} solutions. For each level, we clustered the
feature vectors using the G-means algorithm \citep{Hamerly:PHzvGcYc}, which
functions like a wrapper around k-means using a heuristic that tries to form
clusters with a Gaussian distribution among features to choose a good value
for k. This process produced a clustering of solutions for each level. We
ran it ten times and we report the average and standard deviation
in adjusted Rand index between the CFE and human clusterings.

To examine TRESTLE's ability to cluster {\it RumbleBlocks} solutions, we had it
create a categorization tree for each level, with instances categorized into
the tree in a random order. After all instances had been categorized once, we
shuffled them and categorized them a second time, taking the concepts into
which they were sorted as their labeling. We could have clustered instances
using a single run, but we were interested in trying to maintain parity to the
humans, who were given all the examples at once and allowed to recategorize
them. To model this task we categorized the instances twice to give TRESTLE the
opportunity to form an initial categorization tree from all the examples once
before committing to labels.

Both the human labeling and the CFE clustering produced flat clusterings, so
for comparison purposes we transformed TRESTLE's hierarchical clustering into a
nonhierarchical one. As mentioned earlier, a flat clustering can be produced by
splitting concepts in the tree starting at the root and returning the most
general unsplit concept as a label for each instance. This process can be done
with an arbitrary number of splits; for example, zero splits will return the
root concept as the label for every instance, one split will return the
children of the root as concept labels, and two splits will return a more
refined set of concept labels produced by further splitting one of the
children. This process is similar to the one used to produce flat clusterings
from agglomerative techniques.  Given that the number of splits is
arbitrary, we report the results for the first three splits of the
categorization tree to provide a sense for how the clusterings at different
levels of the hierarchy agree with human clusterings. As with the CFE
evaluation, we repeated this process ten times and we report the average and
standard deviation in adjusted Rand index between the human labelings and each
split of the TRESTLE labelings.

\begin{table}[t!]
\caption{The adjusted Rand index agreement between the model clusterings and human clusterings. We report the average and standard deviation of this measure across ten clustering runs.}
\label{tab:clustering}
\begin{center}
    \small
\begin{tabular}{cccc}
    \hline\\[-2.4ex]
    {\bf Level} & {\bf Model} & {\bf ARI} & {\bf STD}\\
    \hline\\[-1.5ex]
    {\bf Center of Mass} & Trestle (one split)  & 0.37 & 0.03\\
    (n=251)      & Trestle (two splits) & 0.50 & 0.08\\
                   & Trestle (three splits) & {\bf 0.54} & 0.13\\
                   & CFE & 0.51 & 0.08\\[1ex]
    \hline\\[-1.5ex]
    {\bf Symmetry}       & Trestle (one split)  & 0.16 & 0.08\\
    (n=249) & Trestle (two splits) & 0.34 & 0.08\\
                   & Trestle (three splits) & 0.44 & 0.08\\
                   & CFE & {\bf 0.47} & 0.04\\[1ex]
    \hline\\[-1.5ex]
    {\bf Wide Base}      & Trestle (one split)  & {\bf 0.56} & 0.02\\
    (n=254) & Trestle (two splits) & 0.47 & 0.10\\
                   & Trestle (three splits) & 0.41 & 0.05\\
                   & CFE & 0.42 & 0.02\\[1ex]
\hline
\end{tabular}
\end{center}
\end{table}

Table \ref{tab:clustering} compares the two machine clusterings with the human
clustering. For the Center of Mass and Wide Base levels, TRESTLE produced
clusterings with the highest agreement with humans. In the Symmetry level, the
CFE approach produced the most human-like clusterings, but this level had the
greatest number of human clusters (40), so TRESTLE might have benefited from
more splits. In general, TRESTLE seems to better match human behavior than
CFE on this task.

\section{Discussion}

One interpretation of our results is that TRESTLE is a better model of
human behavior while CFE is a better machine learning approach, in that it
achieves greater performance. Both our prediction and clustering results
support this interpretation. First, TRESTLE's predictive accuracy is closer to
human predictive accuracy, while CFE has higher accuracy than both TRESTLE and
humans. Further, TRESTLE's organization of knowledge more closely agrees with
humans than the organization produced using CFE. The two approaches differ, in
part, because of the different motivations underlying their designs. CFE was
originally created as a means of transforming instances in a structured domain
into feature vectors so that other traditional learning algorithms (e.g., CART
or k-means) could function in the space. It was designed primarily for use in
offline settings, such as post-hoc analysis of {\it RumbleBlocks} solutions,
rather than for making online decisions. This is why the algorithm makes use of
{\it exhaustive} grammar generation and computes features based on {\it all}
possible parses of structures, in an attempt to maximize its coverage of the
space despite the additional training costs. TRESTLE, on the other hand, arises
out of existing theory on human categorization and takes a probabilistic
approach to the problem of coverage. Rather than focus on multiple potential
interpretations of a tower it looks at probabilistic similarity with existing
knowledge. Additionally, TRESTLE maintains more psychological plausibility by
avoiding exhaustive processes; for example, it does not seem very plausible
that new players of {\it RumbleBlocks} have a preexisting {\it RumbleBlocks}
grammar, nor that they are visually parsing towers using many grammar rules
(CFE generated about 6,000 for {\it RumbleBlocks}). In summary, our results
support the claim that TRESTLE is a better model of human behavior than CFE
because it is incremental and optimizes for the prediction of any attribute.

Our results also serve as an initial demonstration of TRESTLE's abilities to
operate in domains with nominal, numeric, component, and relational attributes.
While machine learning approaches exist that support each of these
characteristics individually, rarely are they integrated into a single system.
Further, ones that do support numeric, nominal, and relational
attributes, such as FOIL \citep{Quinlan:1990kj}, do not typically
operate incrementally. Thus, we argue that TRESTLE is one of the few systems
that supports all of these capabilities, making it a novel contribution and
particularly well suited for modeling human categorization in a wide range of
settings. For example, there is an active debate about whether blocking or
interleaving similar types of problems is better for student learning
\citep{Carvalho:2013kt}. One hypothesis is that blocking problems supports
students' ability to map structure across similar problems and thus supports
the acquisition of structural knowledge. The contrasting hypothesis is that
interleaving problems better supports students' ability to identify
discriminating features. Previous work with intelligent
tutors has suggested that interleaved instruction is preferable
\citep{Li:2012ex}, but this work first induced a
grammar in batch from all examples (similar to CFE), effectively
pre-blocking learning of common structure. Given this pretrained structural
knowledge, it makes sense that the model would learn best from interleaved
problems because it only needed to identify discriminating features. In
contrast, a model like TRESTLE would be well qualified to explore the issue
of blocked vs. interleaved instruction, as it models the incremental
acquisition of both structural information and discriminating features.

In addition to supporting multiple attribute types, our analysis demonstrates
that TRESTLE supports both supervised and unsupervised learning. We argue that
this makes it a more general model of learning than CFE, which employed
separate machine learning algorithms (i.e., CART for prediction and k-means for
clustering) for the two tasks. In the current work, we evaluated the system on
separate supervised and unsupervised tasks, but TRESTLE should also support
hybrid tasks that involve partial supervision. We claim that this capability
makes TRESTLE ideal for modeling human learning in semi-supervised settings. In
summary, our studies demonstrate the wide range of settings that TRESTLE
supports and provide evidence that it is a better models human behavior than
CFE in the {\it RumbleBlocks} domain. 

\section{Related Work}

We have mentioned previously that the TRESTLE model builds on earlier accounts of
concept formation in the COBWEB family
\citep{Fisher:1987tb,McKusick:2013vm,Gennari:2002tl}. However, we should also
discuss how it relates to these previous systems and other similar models
of human categorization that attempt to accomplish similar goals. One early
account of human categorization was EPAM \citep{Feigenbaum:1984tp}, which
modeled human memorization of nonsense syllables. It learned incrementally
from sequentially presented training examples and took structural information
(i.e., the organizations of letters) into account. Additionally, EPAM organized
its experiences using a discrimination network, which has a tree structure similar
to that used by TRESTLE. COBWEB \citep{Fisher:1987tb} extended this idea to take
into account attribute-value probabilities during concept formation. Further,
COBWEB/3 \citep{McKusick:2013vm} added the ability to handle numeric
information. However, these newer approaches did not support the ability to
handle structural information originally supported by EPAM. 

CLASSIT \citep{Gennari:2002tl} was another variant of COBWEB that supported both
numeric information and component information. Like TRESTLE, it used both
partial matching and flattening steps to handle component information during
categorization. However, CLASSIT lacked support for both nominal or relational
information. LABYRINTH \citep{Thompson:2010wv} was developed to support
component, nominal, and relational information. Like TRESTLE, it utilized a
partial matching step to rename component objects in order to maximize category
utility. During categorization, the system supported components by categorizing
subcomponents in a bottom-up fashion and replacing component values with
nominal concept labels. LABYRINTH also generalized subcomponent labels during
the categorization of higher-level components. This bottom-up categorization
approach let it utilize subcomponent information to improve prediction of 
higher-level attributes, but it could not use higher-level component
information when categorizing subcomponents. In contrast, TRESTLE's flattening
mechanism takes into account the attribute-values of all components in a
holistic way when making predictions. Finally, neither CLASSIT nor LABRYINTH
have been compared to humans or other systems, so it is unclear how 
their performance would compare to that of TRESTLE. 

There also exist other systems that have similar goals, but do not derive from
the COBWEB family. For example, SimStudent \citep{Li:2014hr} shares TRESTLE's
aim of modeling novice learning via a combination of statistical and
structural learning. As in TRESTLE, SimStudent combines both unsupervised
(perceptual representation) and supervised (skill) learning. One key difference
is that TRESTLE is more unified (with a single learning mechanism) and fully
incremental, whereas SimStudent has four learning mechanisms (one for
representations and three for skills), with representation learning running in
batch prior to skill learning. For comparison purposes, CFE is analogous to
SimStudent, in that they use similar grammar learners and
classification/clustering algorithms. To the extent that CFE is a good
approximation of SimStudent, our results suggest that TRESTLE may 
model human learning better than SimStudent in that the latter, like CFE, is
likely to learn faster than humans. More generally, our evidence supports the
hypothesis that an incremental integration of representation and classification
learning better models human behavior than separate, nonincremental,
representation and classification learners. Elements of skill learning that go
beyond classification (e.g., SimStudent's action-sequence learner) are targets
for further research on TRESTLE. 

SAGE \citep{Kuehne:2000ub,McLure:2010wx} is another system that shares the goal
of modeling human concept formation in structured domains. It learns concepts
by matching new instances to existing structures and generalizing matching
concepts to include the instances. If no match is found, then it creates a new
concept to cover the instance. For comparison purposes, SAGE is analogous to a
form of TRESTLE that is nonhierarchical; it maintains only the children of the root
and it cannot split merged concepts because it does not maintain more specific
variants. This difference makes it harder for SAGE to
recover from early misconceptions and biases it towards forming overly
abstract concepts \citep{Kuehne:2000ub}. Thus, in many respects TRESTLE could be
viewed as an extension of SAGE that overcomes these limitations and adds
support for nominal and numerical information. Another key
difference is that SAGE partially matches each instance to every concept,
whereas TRESTLE only partially matches instances to its root 
concept.\footnote{TRESTLE compares the similarity of matched instances to its
concepts during categorization, but it does not perform additional partial
matching.} This difference makes its partial matching more efficient (one match
vs. many matches), but it is unclear how it affects performance. Future
versions of TRESTLE should explore this tradeoff. One strength of SAGE is that
it agrees with data on human behavior \citep{Kuehne:2000ub}. It would be
interesting to explore how the differences between the two models impact their
agreement with observations.

In summary, the literature includes several incremental and nonincremental
models of concept formation. However, each model tends to support only a subset
of attribute types and rarely is their behavior compared with that of other
systems or with humans. TRESTLE is novel in that it unifies facets of previous
incremental approaches by supporting concept formation over nominal, numeric,
structural, and relational information. It also extends systems like SAGE to
support concept revision. Finally, its behavior on {\it RumbleBlocks} compares
favorably to that found in humans.

\section{Conclusion and Future Work}

In this paper, we presented TRESTLE and demonstrated its ability to
incrementally learn concepts from mixed-data environments in both a supervised
and unsupervised fashion. However, our evaluation of the system is preliminary,
and more work is necessary to fully assess its capabilities. In particular, we
should evaluate TRESTLE's ability to perform more flexible prediction tasks
\citep{Fisher:1987tb}, but first, we must determine how to
frame these tasks in structured domains. For example, should we evaluate a
system's ability to predict specific attribute values, or should we assess its
ability to predict entire missing components with multiple attribute values?
This problem is complicated by structure mapping, which makes it
difficult to align component predictions during evaluation. 

In addition to assessing the current system's capabilities, we should integrate
it with methods for planning and executing action. For example, we could extend
TRESTLE to support ``functional'' attributes that have actions as values. This
would let it decide what action to take given an instance. Unlike systems that
separate concept and skill knowledge \citep[e.g.,][]{Langley:2009dt}, this
extension would let TRESTLE use functional information to guide its formation
of concepts and vice versa.

We should also explore the implications of TRESTLE for supporting
designers. Prior work shows that concept formation systems can be
used in this area. For example, \citet{reich1993development} developed
BRIDGER to support bridge design, but his approach did not support component
or relational information. More recently, \citet{Talton:2012wz} demonstrated
how to generate novel designs using grammar patterns induced from examples
provided by designers, but their work did not support
mixed data types or missing attributes. TRESTLE could extend both
approaches to support novel tasks, such as completing partially specified designs
that are described with mixed data types. We would also like to
explore how it could help game designers better understand the
space of player solutions by clustering solutions with common structural
patterns. This application was part of the original motivation for CFE, but our
clustering results suggest that TRESTLE is better suited to the task and may
even provide more information through its hierarchical organization of
concepts.

Finally, we should explore how we can use the system as a computational
model of human learning in educational settings. In particular, we would like
it to stimulate student's learning in intelligent tutoring systems and
educational games. In these contexts, the model could be used to test and
improve different instructional materials before they are given to students and
thus determine the best ordering of activities. Additionally, TRESTLE could
automatically construct domain models, in the form of concept hierarchies,
through training \citep{harpstead:2015}. Whether they contain correct or buggy
knowledge, such models could then be used in intelligent tutoring systems to
provide students with improved context-sensitive feedback and instruction. 

In summary, TRESTLE is a model of incremental concept
formation in structured domains. It carries out supervised learning and
unsupervised clustering over nominal, numeric, component, and relational
information. Experiments show that it can achieve performance that is
comparable to CFE, a previously developed nonincremental approach. Furthermore,
although CFE achieves higher predictive accuracy, TRESTLE better matches human
behavior because it takes key aspects of human categorization into account.
Our studies demonstrated these capabilities and provided an initial
demonstration of how it can be used to model human categorization.
 
\begin{acknowledgements} 

\noindent
We would like to thank the designers of {\it RumbleBlocks} and our colleagues
who conducted the formative evaluation that yielded our data. We would also like
to thank Pat Langley, Victor Hwang, Kelly Rivers, Eliane Stampfer Wiese, and
Douglas Fisher for their feedback and comments on earlier versions of this
work. This research was supported in part by the Department of Education
(\#R305B090023), National Science Foundation (\#SBE-0836012), and the DARPA
ENGAGE research program under ONR Contract Number N00014-12-C-0284. All
opinions expressed in this article are those of the authors and do not
necessarily reflect the position of the sponsoring agency.
\end{acknowledgements} 

\vspace{-0.25in}

{\parindent -10pt\leftskip 10pt\noindent
\bibliographystyle{cogsysapa}
\bibliography{format}

}


\end{document}